\documentclass{article} 
\usepackage{iclr2025_conference,times}

\usepackage{amsmath,amsfonts,bm}

\def\eqref#1{equation~\ref{#1}}

\def\1{\bm{1}}

\DeclareMathAlphabet{\mathsfit}{\encodingdefault}{\sfdefault}{m}{sl}
\SetMathAlphabet{\mathsfit}{bold}{\encodingdefault}{\sfdefault}{bx}{n}

\usepackage{hyperref}
\usepackage{url}
\usepackage{xcolor}     

\usepackage{graphicx}
\usepackage{multirow}
\usepackage{subfigure}
\usepackage{listings}
\title{Rationalization Models for Text-to-SQL}

\author{Gaetano Rossiello, Nhan Pham, Michael Glass, Junkyu Lee, Dharmashankar Subramanian \\
IBM T.J. Watson Research Center\\Yorktown Heights, NY, USA}

\iclrfinalcopy 
\begin{document}

\maketitle

\begin{abstract}

We introduce a framework for generating Chain-of-Thought (CoT) rationales to enhance text-to-SQL model fine-tuning. These rationales consist of intermediate SQL statements and explanations, serving as incremental steps toward constructing the final SQL query. The process begins with manually annotating a small set of examples, which are then used to prompt a large language model in an iterative, dynamic few-shot knowledge distillation procedure from a teacher model. A rationalization model is subsequently trained on the validated decomposed queries, enabling extensive synthetic CoT annotations for text-to-SQL datasets. To evaluate the approach, we fine-tune small language models with and without these rationales on the BIRD dataset. Results indicate that step-by-step query generation improves execution accuracy, especially for moderately and highly complex queries, while also enhancing explainability.
\end{abstract}

\section{Introduction}

%
Recent advances in LLMs have demonstrated breakthroughs in the text-to-SQL task, a long-standing challenge in computer science~\citep{DBLP:journals/tods/HendrixSSS78, DBLP:journals/ker/CopestakeJ90}.
Modern text-to-SQL systems follow a pipeline: schema linking, SQL generation via LLMs, and post-processing (repair, verification, selection).
~\citep{DBLP:journals/corr/abs-2411-08599, DBLP:journals/corr/abs-2410-01943, DBLP:journals/corr/abs-2408-07702, DBLP:journals/corr/abs-2405-16755}.
Nevertheless, the performance of the best-performing systems is still far from human-level performance, especially in enterprise use cases~\citep{DBLP:conf/nips/LiHQYLLWQGHZ0LC23, DBLP:journals/corr/abs-2411-07763}.

In parallel to complex natural language reasoning tasks,
question decomposition approaches have shown to be a promising direction
for complex text-to-SQL tasks~\citep{ DBLP:conf/emnlp/PourrezaR23, DBLP:journals/corr/abs-2307-07306, DBLP:journals/pvldb/GaoWLSQDZ24, DBLP:conf/coling/WangR0LBCYZYSL25}.
Rather than generating the desired SQL in a single pass, decomposition approaches break down 
the input natural language question into a sequence of simpler sub-questions and employ 
the step-by-step reasoning approach. 
The CoT style in-context learning methods often help LLMs to generate more accurate results and intermediate thoughts provide human-readable explanations. 
Moreover, CoT steps are essential for self-improvement~\citep{DBLP:conf/iclr/LightmanKBEBLLS24, deepseekai2025deepseekr1incentivizingreasoningcapability} and 
for enhancing the reasoning capabilities of smaller models by fine-tuning them through knowledge distillation~\citep{DBLP:conf/nips/ZelikmanWMG22, DBLP:conf/emnlp/0001GHW00023, DBLP:conf/acl/MagisterMAMS23, DBLP:conf/acl/LiHLJSL024}.
%

However, publicly available text-to-SQL datasets often lack step-by-step decompositions, 
and manual annotation can be expensive. 
Therefore, we propose a framework for generating rationales in CoT style for the text-to-SQL task, which comes with unique challenges.
We start with existing text-to-SQL training sets, provide a few human annotations, and incorporate dynamic few-shot sample selection to bootstrap valid rationales and fine-tune rationalization models.

Our preliminary evaluation on the BIRD benchmark shows that fine-tuning smaller models with intermediate rationales consistently enhances performance, especially on moderate to challenging problems. Additionally, these fine-tuned models generate clear explanations of the query-building process and offer actionable plans for implementing corrective measures at each step.

\section{Methodology}

\begin{figure}[t]
    \centering
    \subfigure[]{\includegraphics[width=0.45\textwidth]{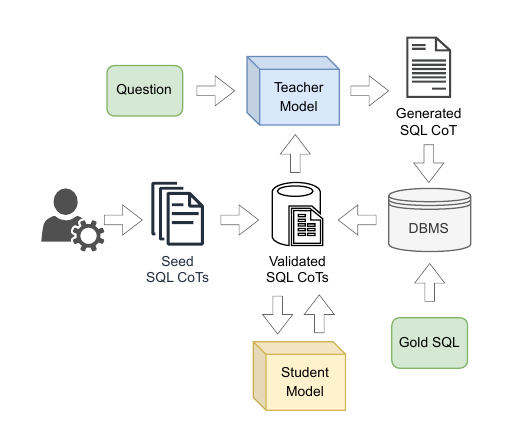}\label{fig:framework}} 
    \subfigure[]{\includegraphics[width=0.5\textwidth]{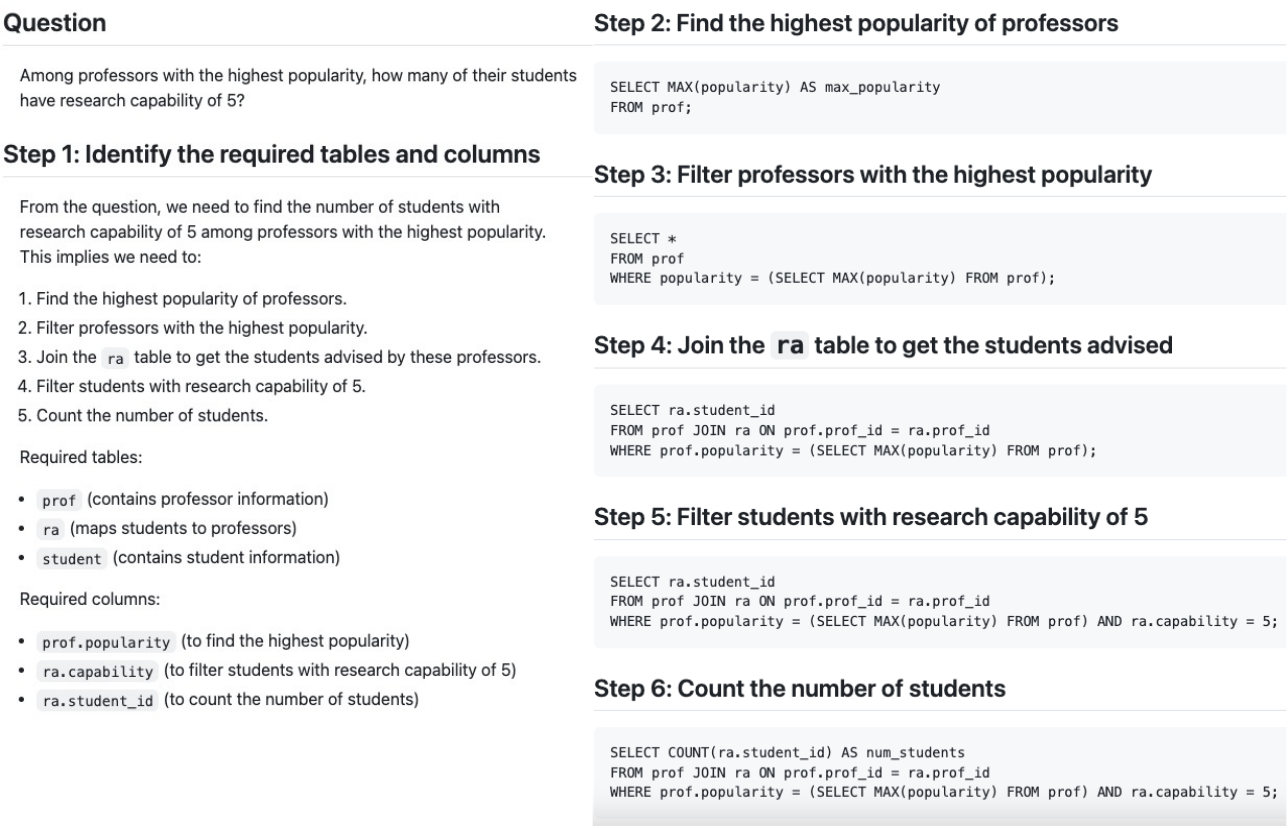}\label{fig:cot_markdown}} 
    \caption{(a) Framework overview (b) Step-by-step SQL building rationale.}
    \label{fig:cot}
\end{figure}


Figure~\ref{fig:framework} provides an overview of the synthetic data generation method. The process involves a subject matter expert who defines the representation of rationales, a text-to-SQL dataset containing (question, gold SQL) pairs, a DBMS for validation, a teacher LLM with few-shot capabilities, and, finally, a smaller student LLM used to fine-tune the rationalization model. 

\subsection{SQL-Building Rationale Representation}
The first stage involves manually annotating a few seed examples with the desired rationale representation.
There are various approaches to generating CoT steps, including manual~\citep{DBLP:conf/nips/Wei0SBIXCLZ22,DBLP:conf/iclr/FuPSCK23}, automatic~\citep{DBLP:conf/nips/KojimaGRMI22,DBLP:journals/tmlr/ChenM0C23}, and semi-automatic~\citep{DBLP:conf/emnlp/ShumDZ23,DBLP:conf/iclr/Lu0CWZRCK23,DBLP:journals/corr/abs-2302-04813} prompting techniques.
Our approach aims to harness the benefits of all three methods by combining them effectively. 
First, we employ automatic prompting with the zero-shot instruction (see Appendix~\ref{appendix_1}) to generate an initial draft of SQL steps.
We then refine this draft by incorporating a preliminary step that outlines the plan and structuring it using Markdown (see Appendix~\ref{appendix_2}).
Markdown facilitates easy parsing of the generated output, allowing for the extraction of individual intermediate SQL statements and enhancing visualization for qualitative assessment.

We develop the rationale by emulating the cognitive process of incrementally constructing an SQL query, joining tables, and adding conditions step-by-step. 
Figure~\ref{fig:cot_markdown} illustrates an example of such a CoT representation. Given a question, the first step involves rephrasing it, outlining the decomposition steps, and identifying the relevant tables and columns while justifying their selection. 
It benefits the user by providing an interpretable overall solution plan and aids the LLM by offering focused context to enhance in-context learning. Then, the SQL-building process continues through a sequence of steps with increasing complexity, providing a textual headline along with the corresponding intermediate, executable SQL. The SQL at the final step represents the solution to the given input question.

\subsection{Knowledge Distillation with Dynamic Few-shot}
Manually curated seed examples are stored in a repository of validated SQL CoTs, to boostrap a dynamic few-shot rationale generation process.
This iterative approach is applied to each (question, gold SQL) pair within a given text-to-SQL training set. For each training instance, the gold SQL serves as a query to rank the most similar examples from the validated SQL CoTs.
To implement the example ranking function, we construct a sparse vector space model, where distinct SQL-reserved keywords define the feature space.
Each validated CoT’s final SQL statement is extracted and transformed into a sparse vector representing SQL keyword frequency.
The gold SQL is projected into the same vector space and used as a query vector to rank SQL CoT examples based on cosine similarity.
Finally, the top-n most similar examples, annotated with SQL-building rationales, are selected as few-shot examples and incorporated into the prompt.
This strategy facilitates the selection of validated CoT examples with similar SQL structures, such as nested queries, aggregation functions, or multi-joins, drawing inspiration from a case-based reasoning approach.

We initiate a dynamic few-shot process by leveraging an LLM as a teacher. The LLM processes a prompt containing the database (DB) schema—filtered by a schema linker~\citep{DBLP:journals/corr/abs-2501-17174}—the user question, a list of top-n similar few-shot examples annotated with CoTs, and an instruction to generate a sequence of reasoning steps. 

A validation component ensures quality by executing the final SQL, from the generated CoT, and the gold SQL, from the training set, against the DB. If the two result sets match exactly, the CoT is labeled as a positive example; otherwise, it is labeled negative. The validated CoTs are iteratively collected into a repository, where they may be selected as few-shot examples in the next iterations. The selection process is guided by the SQL-based similarity function, which clusters the generated CoTs. The dynamic few-shot approach enhances the selection of diverse CoT annotations after each iteration, thereby increasing the proportion of positively validated rationales. The dynamic few-shot generation process iterates over the (question, gold SQL) training pairs until the number of validated SQL CoTs no longer increases.

\subsection{Rationalization Model Fine-tuning}
Despite the dynamic few-shot process with the teacher model significantly enhancing the coverage of validated SQL CoT annotations (see Section~\ref{sec:validation}), some training instances may still contain invalid, synthetically generated rationales.
Following STaR~\citep{DBLP:conf/nips/ZelikmanWMG22}, we distinguish rationales from rationalizations. Rationales are CoTs produced without knowing the target answer, while rationalizations are produced as step-by-step explanations of already known answers.
To achieve complete annotation coverage, we fine-tune a rationalization model and apply it to the remaining training examples that lack correct generated rationales.  Validated CoTs from the previous dynamic few-shot stage provide the training dataset for the rationalization model. The filtered DB schema, the user question, and the gold SQL are the input, and the fully validated SQL-building CoT is the output.

The rationalization model is designed not to generate an SQL solution for a given user request but to produce step-by-step reasoning by incorporating both the question and the correct SQL query into the input prompt. As in the previous stage, the execution of the final SQL query derived from the generated CoT is compared with the execution of the gold-standard SQL. Our findings indicate that a text-to-SQL rationalization model fine-tuned in this manner can help identify incorrect (question, gold SQL) training pairs. In rare cases where the result sets do not match, the provided ground truth may contain inconsistencies that can be easily detected and corrected.

Furthermore, a fine-tuned rationalization model can be utilized to generate SQL-building rationales for other text-to-SQL datasets, even across different domains, without needing to repeat the entire process from scratch. This presents a promising direction for future research.

\section{Experiments}
To evaluate the quality of the generated SQL CoTs, we conducted a controlled experiment using the \texttt{train} and \texttt{dev} sets of the BIRD~\citep{DBLP:conf/nips/LiHQYLLWQGHZ0LC23} dataset. 
This dataset is particularly well-suited to our study, as its \texttt{train} set (containing question-SQL pairs) does not include query decomposition steps. 
Furthermore, the \texttt{dev} set is organized into three main categories -simple, moderate, and challenging SQL statements- enabling a fine-grained, independent evaluation of model performance across different difficulty levels.

We begin by cleaning the training set, removing instances where the gold SQL is not executable, either due to syntax errors, timeouts, or empty result sets. After the cleaning process, we collect 8,807 text-to-SQL training pairs. The original \texttt{dev} set, consisting of 1,534 instances, is retained. We use the open-source Llama 3.1 LLM family~\citep{DBLP:journals/corr/abs-2407-21783}, specifically the 70B version as the teacher for the dynamic few-shot process, and the 8B version as the student to fine-tune both the rationalization model and the text-to-SQL models, with and without CoTs.

\subsection{SQL Rationale Training Coverage}
\label{sec:validation}

\begin{table}[t]
\small
\centering
\bgroup
\def\arraystretch{1.2}
\begin{tabular}{c|c|c}
Stage            & Model         & CoT Train\% coverage \\ \hline
Manual Few-shot  & Llama 3.1 70B & 53.18            \\
Dynamic Few-shot & Llama 3.1 70B & 73.86            \\
Fine-tuning      & Llama 3.1 8B  & \textbf{99.02} \\ \hline           
\end{tabular}
\egroup
\label{table:exp_coverage}
\caption{Percentage of validated SQL CoTs in BIRD training set after each rationalization stage.}
\end{table}

The coverage of validated CoT at each stage is shown in Table~\ref{table:exp_coverage}. After manually annotating two random few-shot examples, the teacher model produces correct CoT for 53.18\% of the training instances. The dynamic few-shot approach increases the coverage by more than 20\% after 13 iterations, alternating between greedy and sampling decoding methods. Finally, applying the fine-tuned rationalization model to the instances with incorrect CoT ensures full training coverage. We manually assessed the remaining 86 (0.98\%) instances and found that, in the majority of cases, the generated CoT was correct, while the provided gold standard contained ambiguities or errors.


\subsection{Impact of SQL Rationales in Fine-tuning Text-to-SQL Models}
\label{sec:impact}
\begin{table}[t]
\small
\centering
\bgroup
\def\arraystretch{1.2}
\begin{tabular}{c|c|c|c|c|c|c|c}
Stage                        & Model                         & Train\%                & Dataset   & Simple & Moderate & Challenge & Total \\ \hline
\multirow{3}{*}{Dyn. Few-shot}    & \multirow{3}{*}{Llama3.1 8B} & \multirow{3}{*}{73.86} & Gold SQL  &   \textbf{72.00}     &   54.53       &     42.76      &   \textbf{63.95}    \\
                             &                               &                        & CoT Short &    70.7    &     \textbf{55.39}     &     46.9      &   63.82    \\
                             &                               &                        & CoT Long  &    69.08    &       52.37   &     \textbf{47.59}      &    61.99   \\ \hline\hline
\multirow{3}{*}{Fine-tuning} & \multirow{3}{*}{Llama3.1 8B}  & \multirow{3}{*}{99.02} & Gold SQL  &    73.08    &     58.06    &    47.92       &  66.17     \\
                             &                               &                        & CoT Short &    72.54    &     59.78     &    52.08       &    66.75   \\
                             &                               &                        & CoT Long  &    \textbf{73.41}    &     \textbf{60.00}     &     \textbf{52.78}      &    \textbf{67.41}   \\\hline
                              
\end{tabular}
\egroup
\label{table:exp_finetuning}
\caption{Execution accuracy comparison on BIRD \texttt{dev} set. \textit{Gold}, \textit{Short CoT}, \textit{Long CoT} refer to the accuracy achieved by the models fine-tuned on the original gold standard SQLs, the shortest CoT and the longest CoT, respectively.}
\end{table}
Due to the iterative nature of dynamic few-shot learning, training instance pairs can generate multiple valid synthetic CoTs. We created six different training sets for the text-to-SQL task, split into two groups based on the number of instances. These sets use the same input prompts but differ in their targets: the gold SQL without CoT, which serves as a baseline, and selections of the shortest/longest CoTs for each train instance. The goal is to evaluate the impact on the execution accuracy of Llama 3.1 8B, fine-tuned with and without query decomposition steps, both before and after the application of the rationalization model. The results of the fine-tuned models applied to the BIRD development set are presented in Table~\ref{table:exp_finetuning}. Using 73.68\% of the training set, after the dynamic few-shot process, a model fine-tuned without CoT demonstrates overall better performance, particularly on the simple questions, which are more frequent. However, the model fine-tuned with the short/long CoT demonstrates a greater ability to handle moderate and challenging questions, respectively. After applying the rationalization model with full coverage of the training set, the text-to-SQL model, fine-tuned on the longest CoT, outperforms in all three categories, improving overall performance by +1.24\%, with a notable increase of +4.86\% on the challenging queries.

\section{Conclusion}
In this work, we presented a framework designed to facilitate the annotation of text-to-SQL datasets with CoT rationales, which consist of intermediate SQL-building steps. The process requires minimal manual annotations, using only a few examples to define the rationale representation. A teacher model is then employed to distill a large volume of validated synthetic CoTs, which are used to bootstrap a rationalization model that enables full training coverage with CoT. We introduced a sample selection method based on SQL similarity, supporting a dynamic few-shot approach. Our comparison demonstrates that long CoTs enhance the accuracy of text-to-SQL tasks while providing an interpretable, user-friendly output that clarifies the reasoning process behind query construction.


\bibliography{iclr2025_conference}

\begin{thebibliography}{27}
\providecommand{\natexlab}[1]{#1}
\providecommand{\url}[1]{\texttt{#1}}
\expandafter\ifx\csname urlstyle\endcsname\relax
  \providecommand{\doi}[1]{doi: #1}\else
  \providecommand{\doi}{doi: \begingroup \urlstyle{rm}\Url}\fi

\bibitem[Chen et~al.(2023)Chen, Ma, Wang, and
  Cohen]{DBLP:journals/tmlr/ChenM0C23}
Wenhu Chen, Xueguang Ma, Xinyi Wang, and William~W. Cohen.
\newblock Program of thoughts prompting: Disentangling computation from
  reasoning for numerical reasoning tasks.
\newblock \emph{Trans. Mach. Learn. Res.}, 2023, 2023.

\bibitem[Copestake \& Jones(1990)Copestake and
  Jones]{DBLP:journals/ker/CopestakeJ90}
Ann~A. Copestake and Karen~Sparck Jones.
\newblock Natural language interfaces to databases.
\newblock \emph{Knowl. Eng. Rev.}, 5\penalty0 (4):\penalty0 225--249, 1990.

\bibitem[DeepSeek-AI et~al.(2025)DeepSeek-AI, Guo, Yang, Zhang, Song, Zhang,
  Xu, Zhu, Ma, Wang, Bi, Zhang, Yu, Wu, Wu, Gou, Shao, Li, Gao, Liu, Xue, Wang,
  Wu, Feng, Lu, Zhao, Deng, Zhang, Ruan, Dai, Chen, Ji, Li, Lin, Dai, Luo, Hao,
  Chen, Li, Zhang, Bao, Xu, Wang, Ding, Xin, Gao, Qu, Li, Guo, Li, Wang, Chen,
  Yuan, Qiu, Li, Cai, Ni, Liang, Chen, Dong, Hu, Gao, Guan, Huang, Yu, Wang,
  Zhang, Zhao, Wang, Zhang, Xu, Xia, Zhang, Zhang, Tang, Li, Wang, Li, Tian,
  Huang, Zhang, Wang, Chen, Du, Ge, Zhang, Pan, Wang, Chen, Jin, Chen, Lu,
  Zhou, Chen, Ye, Wang, Yu, Zhou, Pan, Li, Zhou, Wu, Ye, Yun, Pei, Sun, Wang,
  Zeng, Zhao, Liu, Liang, Gao, Yu, Zhang, Xiao, An, Liu, Wang, Chen, Nie,
  Cheng, Liu, Xie, Liu, Yang, Li, Su, Lin, Li, Jin, Shen, Chen, Sun, Wang,
  Song, Zhou, Wang, Shan, Li, Wang, Wei, Zhang, Xu, Li, Zhao, Sun, Wang, Yu,
  Zhang, Shi, Xiong, He, Piao, Wang, Tan, Ma, Liu, Guo, Ou, Wang, Gong, Zou,
  He, Xiong, Luo, You, Liu, Zhou, Zhu, Xu, Huang, Li, Zheng, Zhu, Ma, Tang,
  Zha, Yan, Ren, Ren, Sha, Fu, Xu, Xie, Zhang, Hao, Ma, Yan, Wu, Gu, Zhu, Liu,
  Li, Xie, Song, Pan, Huang, Xu, Zhang, and
  Zhang]{deepseekai2025deepseekr1incentivizingreasoningcapability}
DeepSeek-AI, Daya Guo, Dejian Yang, Haowei Zhang, Junxiao Song, Ruoyu Zhang,
  Runxin Xu, Qihao Zhu, Shirong Ma, Peiyi Wang, Xiao Bi, Xiaokang Zhang,
  Xingkai Yu, Yu~Wu, Z.~F. Wu, Zhibin Gou, Zhihong Shao, Zhuoshu Li, Ziyi Gao,
  Aixin Liu, Bing Xue, Bingxuan Wang, Bochao Wu, Bei Feng, Chengda Lu,
  Chenggang Zhao, Chengqi Deng, Chenyu Zhang, Chong Ruan, Damai Dai, Deli Chen,
  Dongjie Ji, Erhang Li, Fangyun Lin, Fucong Dai, Fuli Luo, Guangbo Hao,
  Guanting Chen, Guowei Li, H.~Zhang, Han Bao, Hanwei Xu, Haocheng Wang,
  Honghui Ding, Huajian Xin, Huazuo Gao, Hui Qu, Hui Li, Jianzhong Guo, Jiashi
  Li, Jiawei Wang, Jingchang Chen, Jingyang Yuan, Junjie Qiu, Junlong Li, J.~L.
  Cai, Jiaqi Ni, Jian Liang, Jin Chen, Kai Dong, Kai Hu, Kaige Gao, Kang Guan,
  Kexin Huang, Kuai Yu, Lean Wang, Lecong Zhang, Liang Zhao, Litong Wang, Liyue
  Zhang, Lei Xu, Leyi Xia, Mingchuan Zhang, Minghua Zhang, Minghui Tang, Meng
  Li, Miaojun Wang, Mingming Li, Ning Tian, Panpan Huang, Peng Zhang, Qiancheng
  Wang, Qinyu Chen, Qiushi Du, Ruiqi Ge, Ruisong Zhang, Ruizhe Pan, Runji Wang,
  R.~J. Chen, R.~L. Jin, Ruyi Chen, Shanghao Lu, Shangyan Zhou, Shanhuang Chen,
  Shengfeng Ye, Shiyu Wang, Shuiping Yu, Shunfeng Zhou, Shuting Pan, S.~S. Li,
  Shuang Zhou, Shaoqing Wu, Shengfeng Ye, Tao Yun, Tian Pei, Tianyu Sun,
  T.~Wang, Wangding Zeng, Wanjia Zhao, Wen Liu, Wenfeng Liang, Wenjun Gao,
  Wenqin Yu, Wentao Zhang, W.~L. Xiao, Wei An, Xiaodong Liu, Xiaohan Wang,
  Xiaokang Chen, Xiaotao Nie, Xin Cheng, Xin Liu, Xin Xie, Xingchao Liu, Xinyu
  Yang, Xinyuan Li, Xuecheng Su, Xuheng Lin, X.~Q. Li, Xiangyue Jin, Xiaojin
  Shen, Xiaosha Chen, Xiaowen Sun, Xiaoxiang Wang, Xinnan Song, Xinyi Zhou,
  Xianzu Wang, Xinxia Shan, Y.~K. Li, Y.~Q. Wang, Y.~X. Wei, Yang Zhang,
  Yanhong Xu, Yao Li, Yao Zhao, Yaofeng Sun, Yaohui Wang, Yi~Yu, Yichao Zhang,
  Yifan Shi, Yiliang Xiong, Ying He, Yishi Piao, Yisong Wang, Yixuan Tan,
  Yiyang Ma, Yiyuan Liu, Yongqiang Guo, Yuan Ou, Yuduan Wang, Yue Gong, Yuheng
  Zou, Yujia He, Yunfan Xiong, Yuxiang Luo, Yuxiang You, Yuxuan Liu, Yuyang
  Zhou, Y.~X. Zhu, Yanhong Xu, Yanping Huang, Yaohui Li, Yi~Zheng, Yuchen Zhu,
  Yunxian Ma, Ying Tang, Yukun Zha, Yuting Yan, Z.~Z. Ren, Zehui Ren, Zhangli
  Sha, Zhe Fu, Zhean Xu, Zhenda Xie, Zhengyan Zhang, Zhewen Hao, Zhicheng Ma,
  Zhigang Yan, Zhiyu Wu, Zihui Gu, Zijia Zhu, Zijun Liu, Zilin Li, Ziwei Xie,
  Ziyang Song, Zizheng Pan, Zhen Huang, Zhipeng Xu, Zhongyu Zhang, and Zhen
  Zhang.
\newblock Deepseek-r1: Incentivizing reasoning capability in llms via
  reinforcement learning, 2025.

\bibitem[Dong et~al.(2023)Dong, Zhang, Ge, Mao, Gao, Chen, Lin, and
  Lou]{DBLP:journals/corr/abs-2307-07306}
Xuemei Dong, Chao Zhang, Yuhang Ge, Yuren Mao, Yunjun Gao, Lu~Chen, Jinshu Lin,
  and Dongfang Lou.
\newblock {C3:} zero-shot text-to-sql with chatgpt.
\newblock \emph{CoRR}, abs/2307.07306, 2023.

\bibitem[Dubey et~al.(2024)Dubey, Jauhri, Pandey, Kadian, Al{-}Dahle, Letman,
  Mathur, Schelten, Yang, Fan, Goyal, Hartshorn, Yang, Mitra, Sravankumar,
  Korenev, Hinsvark, Rao, Zhang, Rodriguez, Gregerson, Spataru, Rozi{\`{e}}re,
  Biron, Tang, Chern, Caucheteux, Nayak, Bi, Marra, McConnell, Keller, Touret,
  Wu, Wong, Ferrer, Nikolaidis, Allonsius, Song, Pintz, Livshits, Esiobu,
  Choudhary, Mahajan, Garcia{-}Olano, Perino, Hupkes, Lakomkin, AlBadawy,
  Lobanova, Dinan, Smith, Radenovic, Zhang, Synnaeve, Lee, Anderson, Nail,
  Mialon, Pang, Cucurell, Nguyen, Korevaar, Xu, Touvron, Zarov, Ibarra,
  Kloumann, Misra, Evtimov, Copet, Lee, Geffert, Vranes, Park, Mahadeokar,
  Shah, van~der Linde, Billock, Hong, Lee, Fu, Chi, Huang, Liu, Wang, Yu,
  Bitton, Spisak, Park, Rocca, Johnstun, Saxe, Jia, Alwala, Upasani, Plawiak,
  Li, Heafield, Stone, and et~al.]{DBLP:journals/corr/abs-2407-21783}
Abhimanyu Dubey, Abhinav Jauhri, Abhinav Pandey, Abhishek Kadian, Ahmad
  Al{-}Dahle, Aiesha Letman, Akhil Mathur, Alan Schelten, Amy Yang, Angela Fan,
  Anirudh Goyal, Anthony Hartshorn, Aobo Yang, Archi Mitra, Archie Sravankumar,
  Artem Korenev, Arthur Hinsvark, Arun Rao, Aston Zhang, Aur{\'{e}}lien
  Rodriguez, Austen Gregerson, Ava Spataru, Baptiste Rozi{\`{e}}re, Bethany
  Biron, Binh Tang, Bobbie Chern, Charlotte Caucheteux, Chaya Nayak, Chloe Bi,
  Chris Marra, Chris McConnell, Christian Keller, Christophe Touret, Chunyang
  Wu, Corinne Wong, Cristian~Canton Ferrer, Cyrus Nikolaidis, Damien Allonsius,
  Daniel Song, Danielle Pintz, Danny Livshits, David Esiobu, Dhruv Choudhary,
  Dhruv Mahajan, Diego Garcia{-}Olano, Diego Perino, Dieuwke Hupkes, Egor
  Lakomkin, Ehab AlBadawy, Elina Lobanova, Emily Dinan, Eric~Michael Smith,
  Filip Radenovic, Frank Zhang, Gabriel Synnaeve, Gabrielle Lee, Georgia~Lewis
  Anderson, Graeme Nail, Gr{\'{e}}goire Mialon, Guan Pang, Guillem Cucurell,
  Hailey Nguyen, Hannah Korevaar, Hu~Xu, Hugo Touvron, Iliyan Zarov,
  Imanol~Arrieta Ibarra, Isabel~M. Kloumann, Ishan Misra, Ivan Evtimov, Jade
  Copet, Jaewon Lee, Jan Geffert, Jana Vranes, Jason Park, Jay Mahadeokar, Jeet
  Shah, Jelmer van~der Linde, Jennifer Billock, Jenny Hong, Jenya Lee, Jeremy
  Fu, Jianfeng Chi, Jianyu Huang, Jiawen Liu, Jie Wang, Jiecao Yu, Joanna
  Bitton, Joe Spisak, Jongsoo Park, Joseph Rocca, Joshua Johnstun, Joshua Saxe,
  Junteng Jia, Kalyan~Vasuden Alwala, Kartikeya Upasani, Kate Plawiak, Ke~Li,
  Kenneth Heafield, Kevin Stone, and et~al.
\newblock The llama 3 herd of models.
\newblock \emph{CoRR}, abs/2407.21783, 2024.

\bibitem[Fu et~al.(2023)Fu, Peng, Sabharwal, Clark, and
  Khot]{DBLP:conf/iclr/FuPSCK23}
Yao Fu, Hao Peng, Ashish Sabharwal, Peter Clark, and Tushar Khot.
\newblock Complexity-based prompting for multi-step reasoning.
\newblock In \emph{{ICLR}}. OpenReview.net, 2023.

\bibitem[Gao et~al.(2024{\natexlab{a}})Gao, Wang, Li, Sun, Qian, Ding, and
  Zhou]{DBLP:journals/pvldb/GaoWLSQDZ24}
Dawei Gao, Haibin Wang, Yaliang Li, Xiuyu Sun, Yichen Qian, Bolin Ding, and
  Jingren Zhou.
\newblock Text-to-sql empowered by large language models: {A} benchmark
  evaluation.
\newblock \emph{Proc. {VLDB} Endow.}, 17\penalty0 (5):\penalty0 1132--1145,
  2024{\natexlab{a}}.

\bibitem[Gao et~al.(2024{\natexlab{b}})Gao, Liu, Li, Shi, Zhu, Wang, Li, Li,
  Hong, Luo, Gao, Mou, and Li]{DBLP:journals/corr/abs-2411-08599}
Yingqi Gao, Yifu Liu, Xiaoxia Li, Xiaorong Shi, Yin Zhu, Yiming Wang, Shiqi Li,
  Wei Li, Yuntao Hong, Zhiling Luo, Jinyang Gao, Liyu Mou, and Yu~Li.
\newblock Xiyan-sql: {A} multi-generator ensemble framework for text-to-sql.
\newblock \emph{CoRR}, abs/2411.08599, 2024{\natexlab{b}}.

\bibitem[Glass et~al.(2025)Glass, Eyceoz, Subramanian, Rossiello, Vu, and
  Gliozzo]{DBLP:journals/corr/abs-2501-17174}
Michael~R. Glass, Mustafa Eyceoz, Dharmashankar Subramanian, Gaetano Rossiello,
  Long Vu, and Alfio Gliozzo.
\newblock Extractive schema linking for text-to-sql.
\newblock \emph{CoRR}, abs/2501.17174, 2025.

\bibitem[Hendrix et~al.(1978)Hendrix, Sacerdoti, Sagalowicz, and
  Slocum]{DBLP:journals/tods/HendrixSSS78}
Gary~G. Hendrix, Earl~D. Sacerdoti, Daniel Sagalowicz, and Jonathan Slocum.
\newblock Developing a natural language interface to complex data.
\newblock \emph{{ACM} Trans. Database Syst.}, 3\penalty0 (2):\penalty0
  105--147, 1978.

\bibitem[Huang et~al.(2023)Huang, Gu, Hou, Wu, Wang, Yu, and
  Han]{DBLP:conf/emnlp/0001GHW00023}
Jiaxin Huang, Shixiang Gu, Le~Hou, Yuexin Wu, Xuezhi Wang, Hongkun Yu, and
  Jiawei Han.
\newblock Large language models can self-improve.
\newblock In \emph{{EMNLP}}, pp.\  1051--1068. Association for Computational
  Linguistics, 2023.

\bibitem[Kojima et~al.(2022)Kojima, Gu, Reid, Matsuo, and
  Iwasawa]{DBLP:conf/nips/KojimaGRMI22}
Takeshi Kojima, Shixiang~Shane Gu, Machel Reid, Yutaka Matsuo, and Yusuke
  Iwasawa.
\newblock Large language models are zero-shot reasoners.
\newblock In \emph{NeurIPS}, 2022.

\bibitem[Lei et~al.(2024)Lei, Chen, Ye, Cao, Shin, Su, Suo, Gao, Hu, Yin,
  Zhong, Xiong, Sun, Liu, Wang, and Yu]{DBLP:journals/corr/abs-2411-07763}
Fangyu Lei, Jixuan Chen, Yuxiao Ye, Ruisheng Cao, Dongchan Shin, Hongjin Su,
  Zhaoqing Suo, Hongcheng Gao, Wenjing Hu, Pengcheng Yin, Victor Zhong, Caiming
  Xiong, Ruoxi Sun, Qian Liu, Sida~I. Wang, and Tao Yu.
\newblock Spider 2.0: Evaluating language models on real-world enterprise
  text-to-sql workflows.
\newblock \emph{CoRR}, abs/2411.07763, 2024.

\bibitem[Li et~al.(2023)Li, Hui, Qu, Yang, Li, Li, Wang, Qin, Geng, Huo, Zhou,
  Ma, Li, Chang, Huang, Cheng, and Li]{DBLP:conf/nips/LiHQYLLWQGHZ0LC23}
Jinyang Li, Binyuan Hui, Ge~Qu, Jiaxi Yang, Binhua Li, Bowen Li, Bailin Wang,
  Bowen Qin, Ruiying Geng, Nan Huo, Xuanhe Zhou, Chenhao Ma, Guoliang Li,
  Kevin~Chen{-}Chuan Chang, Fei Huang, Reynold Cheng, and Yongbin Li.
\newblock Can {LLM} already serve as {A} database interface? {A} big bench for
  large-scale database grounded text-to-sqls.
\newblock In \emph{NeurIPS}, 2023.

\bibitem[Li et~al.(2024)Li, He, Lei, JunYang, Su, Liu, and
  Zhao]{DBLP:conf/acl/LiHLJSL024}
Xiang Li, Shizhu He, Fangyu Lei, JunYang JunYang, Tianhuang Su, Kang Liu, and
  Jun Zhao.
\newblock Teaching small language models to reason for knowledge-intensive
  multi-hop question answering.
\newblock In \emph{{ACL} (Findings)}, pp.\  7804--7816. Association for
  Computational Linguistics, 2024.

\bibitem[Lightman et~al.(2024)Lightman, Kosaraju, Burda, Edwards, Baker, Lee,
  Leike, Schulman, Sutskever, and Cobbe]{DBLP:conf/iclr/LightmanKBEBLLS24}
Hunter Lightman, Vineet Kosaraju, Yuri Burda, Harrison Edwards, Bowen Baker,
  Teddy Lee, Jan Leike, John Schulman, Ilya Sutskever, and Karl Cobbe.
\newblock Let's verify step by step.
\newblock In \emph{{ICLR}}. OpenReview.net, 2024.

\bibitem[Lu et~al.(2023)Lu, Qiu, Chang, Wu, Zhu, Rajpurohit, Clark, and
  Kalyan]{DBLP:conf/iclr/Lu0CWZRCK23}
Pan Lu, Liang Qiu, Kai{-}Wei Chang, Ying~Nian Wu, Song{-}Chun Zhu, Tanmay
  Rajpurohit, Peter Clark, and Ashwin Kalyan.
\newblock Dynamic prompt learning via policy gradient for semi-structured
  mathematical reasoning.
\newblock In \emph{{ICLR}}. OpenReview.net, 2023.

\bibitem[Maamari et~al.(2024)Maamari, Abubaker, Jaroslawicz, and
  Mhedhbi]{DBLP:journals/corr/abs-2408-07702}
Karime Maamari, Fadhil Abubaker, Daniel Jaroslawicz, and Amine Mhedhbi.
\newblock The death of schema linking? text-to-sql in the age of well-reasoned
  language models.
\newblock \emph{CoRR}, abs/2408.07702, 2024.

\bibitem[Magister et~al.(2023)Magister, Mallinson, Ad{\'{a}}mek, Malmi, and
  Severyn]{DBLP:conf/acl/MagisterMAMS23}
Lucie~Charlotte Magister, Jonathan Mallinson, Jakub Ad{\'{a}}mek, Eric Malmi,
  and Aliaksei Severyn.
\newblock Teaching small language models to reason.
\newblock In \emph{{ACL} {(2)}}, pp.\  1773--1781. Association for
  Computational Linguistics, 2023.

\bibitem[Pourreza \& Rafiei(2023)Pourreza and
  Rafiei]{DBLP:conf/emnlp/PourrezaR23}
Mohammadreza Pourreza and Davood Rafiei.
\newblock Evaluating cross-domain text-to-sql models and benchmarks.
\newblock In \emph{{EMNLP}}, pp.\  1601--1611. Association for Computational
  Linguistics, 2023.

\bibitem[Pourreza et~al.(2024)Pourreza, Li, Sun, Chung, Talaei, Kakkar, Gan,
  Saberi, Ozcan, and Arik]{DBLP:journals/corr/abs-2410-01943}
Mohammadreza Pourreza, Hailong Li, Ruoxi Sun, Yeounoh Chung, Shayan Talaei,
  Gaurav~Tarlok Kakkar, Yu~Gan, Amin Saberi, Fatma Ozcan, and Sercan~{\"{O}}.
  Arik.
\newblock {CHASE-SQL:} multi-path reasoning and preference optimized candidate
  selection in text-to-sql.
\newblock \emph{CoRR}, abs/2410.01943, 2024.

\bibitem[Shum et~al.(2023)Shum, Diao, and Zhang]{DBLP:conf/emnlp/ShumDZ23}
Kashun Shum, Shizhe Diao, and Tong Zhang.
\newblock Automatic prompt augmentation and selection with chain-of-thought
  from labeled data.
\newblock In \emph{{EMNLP} (Findings)}, pp.\  12113--12139. Association for
  Computational Linguistics, 2023.

\bibitem[Talaei et~al.(2024)Talaei, Pourreza, Chang, Mirhoseini, and
  Saberi]{DBLP:journals/corr/abs-2405-16755}
Shayan Talaei, Mohammadreza Pourreza, Yu{-}Chen Chang, Azalia Mirhoseini, and
  Amin Saberi.
\newblock {CHESS:} contextual harnessing for efficient {SQL} synthesis.
\newblock \emph{CoRR}, abs/2405.16755, 2024.

\bibitem[Wang et~al.(2025)Wang, Ren, Yang, Liang, Bai, Chai, Yan, Zhang, Yin,
  Sun, and Li]{DBLP:conf/coling/WangR0LBCYZYSL25}
Bing Wang, Changyu Ren, Jian Yang, Xinnian Liang, Jiaqi Bai, Linzheng Chai,
  Zhao Yan, Qian{-}Wen Zhang, Di~Yin, Xing Sun, and Zhoujun Li.
\newblock {MAC-SQL:} {A} multi-agent collaborative framework for text-to-sql.
\newblock In \emph{{COLING}}, pp.\  540--557. Association for Computational
  Linguistics, 2025.

\bibitem[Wei et~al.(2022)Wei, Wang, Schuurmans, Bosma, Ichter, Xia, Chi, Le,
  and Zhou]{DBLP:conf/nips/Wei0SBIXCLZ22}
Jason Wei, Xuezhi Wang, Dale Schuurmans, Maarten Bosma, Brian Ichter, Fei Xia,
  Ed~H. Chi, Quoc~V. Le, and Denny Zhou.
\newblock Chain-of-thought prompting elicits reasoning in large language
  models.
\newblock In \emph{NeurIPS}, 2022.

\bibitem[Ye \& Durrett(2023)Ye and Durrett]{DBLP:journals/corr/abs-2302-04813}
Xi~Ye and Greg Durrett.
\newblock Explanation selection using unlabeled data for in-context learning.
\newblock \emph{CoRR}, abs/2302.04813, 2023.

\bibitem[Zelikman et~al.(2022)Zelikman, Wu, Mu, and
  Goodman]{DBLP:conf/nips/ZelikmanWMG22}
Eric Zelikman, Yuhuai Wu, Jesse Mu, and Noah~D. Goodman.
\newblock Star: Bootstrapping reasoning with reasoning.
\newblock In \emph{NeurIPS}, 2022.

\end{thebibliography}
\bibliographystyle{iclr2025_conference}

\newpage
\appendix
\section{Appendix}
\subsection{Text-to-SQL Prompt Representation (Source)}
\label{appendix_1}
\begin{lstlisting}
[SCHEMA]
CREATE TABLE prof (
 prof_id INTEGER PRIMARY KEY, -- unique id for professors
 first_name TEXT, -- the first name of the professor
 last_name TEXT, -- the last name of the professor
 popularity INTEGER, -- popularity of the professor
);

prof.prof_id
prof.first_name: 'Bernhard', 'Hattie', 'Mateo'
prof.last_name: 'Conkay', 'Cunningham', 'Ewenson'
prof.popularity: 2, 3

CREATE TABLE registration (
 student_id INTEGER, -- the id of students
 FOREIGN KEY(student_id) REFERENCES student(student_id)
);

registration.student_id

CREATE TABLE ra (
 student_id INTEGER, -- the id numbe representing each student
 capability INTEGER, -- the capability of student on research
-- (Evaluated by the professor)
 prof_id INTEGER, -- professor who advises this student
 FOREIGN KEY(student_id) REFERENCES student(student_id)
 FOREIGN KEY(prof_id) REFERENCES prof(prof_id)
);

ra.student_id
ra.capability: 2, 4, 5
ra.prof_id

CREATE TABLE student (
 student_id INTEGER PRIMARY KEY, -- the unique id to identify students
 f_name TEXT, -- the first name of the student
);

student.student_id
student.f_name: 'Ahmed', 'Alvera', 'Amerigo'

selected: prof.popularity, prof.prof_id, ra.student_id, None, ra.prof_id, ra.capability, student.student_id, prof.last_name, student.f_name, prof.first_name
join: ra.prof_id, prof.prof_id, None, student.student_id, ra.student_id, registration.student_id, ra.capability, prof.popularity
condition: ra.capability, prof.popularity, None, prof.prof_id, ra.prof_id, ra.student_id, student.student_id, prof.last_name
order: prof.popularity, None, ra.student_id, ra.prof_id, prof.prof_id, student.student_id, ra.capability
group: None, ra.prof_id, prof.prof_id, prof.popularity, ra.student_id, ra.capability, prof.last_name

Comparisons:
ra.capability
3 (many), 5 (many), 4 (many)
ra.capability = 5
5 (many)

Note:
highest popularity refers to MAX(popularity)
research capability refers to capability
capability = 5
[/SCHEMA]

[QUESTION]
Among professors with the highest popularity, how many of their students have research capability of 5?
[/QUESTION]

Given the above [SCHEMA] of a database and a question [QUESTION], translate the question into a valid SQLite statement. Decompose the SQL in increasingly complex building blocks. Explain each step of the SQL building process thinking step by step. Format the output using the Markdown language.
\end{lstlisting}

\subsection{SQL Rationale Representation (Target)}
\label{appendix_2}
\begin{lstlisting}
**Step 1: Identify the required tables and columns**
--

From the question, we need to find the number of students with research capability of 5 among professors with the highest popularity. This implies we need to:

1. Find the highest popularity of professors.
2. Filter professors with the highest popularity.
3. Join the `ra` table to get the students advised by these professors.
4. Filter students with research capability of 5.
5. Count the number of students.

Required tables:
* `prof` (contains professor information)
* `ra` (maps students to professors)
* `student` (contains student information)

Required columns:
* `prof.popularity` (to find the highest popularity)
* `ra.capability` (to filter students with research capability of 5)
* `ra.student_id` (to count the number of students)

**Step 2: Find the highest popularity of professors**
--

```sql
SELECT MAX(popularity) AS max_popularity
FROM prof;
```

**Step 3: Filter professors with the highest popularity**
--

```sql
SELECT *
FROM prof
WHERE popularity = (SELECT MAX(popularity) FROM prof);
```

**Step 4: Join the `ra` table to get the students advised by these professors**
--

```sql
SELECT ra.student_id
FROM prof JOIN ra ON prof.prof_id = ra.prof_id
WHERE prof.popularity = (SELECT MAX(popularity) FROM prof);
```

**Step 5: Filter students with research capability of 5**
--

```sql
SELECT ra.student_id
FROM prof JOIN ra ON prof.prof_id = ra.prof_id
WHERE prof.popularity = (SELECT MAX(popularity) FROM prof) AND ra.capability = 5;
```

**Step 6: Count the number of students**
--

```sql
SELECT COUNT(ra.student_id) AS num_students
FROM prof JOIN ra ON prof.prof_id = ra.prof_id
WHERE prof.popularity = (SELECT MAX(popularity) FROM prof) AND ra.capability = 5;
```

This is the final SQL statement that answers the question.
\end{lstlisting}

\end{document}